\documentclass[11pt,a4paper]{article}
\usepackage[hyperref]{emnlp-ijcnlp-2019}
\usepackage{times}
\usepackage{latexsym}
\usepackage{enumitem}
\usepackage{graphicx}
\usepackage{url}
\usepackage{amssymb}
\usepackage{amsmath}
\usepackage{appendix}
\usepackage{multirow}
\usepackage{booktabs}
\usepackage{subfig}
\usepackage{booktabs}
\usepackage{xcolor}%
\usepackage{todonotes}
\usepackage[T1]{fontenc}  %
\usepackage{comment}
\usepackage{array}
\usepackage{algorithm}
\usepackage{algpseudocode}
\usepackage{mathtools}
\usepackage{caption}

\usepackage{etoolbox}
\newtoggle{draft}
\definecolor{Process Blue}{RGB}{0, 176, 240}
\definecolor{Dark Teal}{RGB}{0, 75, 80}
\definecolor{Fern Green}{RGB}{57, 150, 33}
\definecolor{Red}{RGB}{100, 0, 0}

\toggletrue{draft}
\iftoggle{draft}{
\newcommand{\hh}[1]{\textcolor{Process Blue}{[#1 \textsc{-}]}}

\newcommand{\gn}[1]{\textcolor{magenta}{[#1 \textsc{--}]}}
\newcommand{\zh}[1]{\textcolor{Fern Green}{[#1 \textsc{-}]}}
}{
\newcommand{\hh}[1]{}
\newcommand{\gn}[1]{}
\newcommand{\zh}[1]{}
}

\aclfinalcopy %

\newcommand{\given}{\,|\,}  %
\newcommand{\unpack}{\hspace{-0.3em}:\!}
\makeatletter
\let\OldStatex\Statex
\renewcommand{\Statex}[1][3]{%
  \setlength\@tempdima{\algorithmicindent}%
  \OldStatex\hskip\dimexpr#1\@tempdima\relax}
\makeatother

\title{Latent Relation Language Models}

\author{Hiroaki Hayashi$^{\dagger*}$, Zecong Hu$^{\dagger*}$, Chenyan Xiong$^{\ddagger}$, Graham Neubig$^{\dagger}$ \\
$^\dagger$Carnegie Mellon University, $^\ddagger$Microsoft Research AI}

\date{}

\begin{document}
\maketitle
\begin{abstract}
\renewcommand*{\thefootnote}{\fnsymbol{footnote}}
\setcounter{footnote}{1}
In this paper, we propose Latent Relation Language Models (LRLMs), a class of language models that parameterizes the joint distribution over the words in a document and the entities that occur therein via knowledge graph relations.
This model has a number of attractive properties: it not only improves language modeling performance, but is also able to annotate the posterior probability of entity spans for a given text through relations.
Experiments demonstrate empirical improvements over both a word-based baseline language model and a previous approach that incorporates knowledge graph information.
Qualitative analysis further demonstrates the proposed model's ability to learn to predict appropriate relations in context.%
\footnotetext{Equal contribution.}
\end{abstract}

\setcounter{footnote}{1}

\section{Introduction}
\label{sec:intro}

Language models (LMs) calculate the probability $P(X)$ of textual data $X$, and are a core model class of interest to NLP. LMs are used as testbeds for evaluation of generative models of text, and have applications such as rescoring of upstream language generation inputs~\cite{sundermeyer2012lstm}, grammatical error correction~\cite{felice2014grammatical}, or pre-training of sentence representations \cite{dai2015semi,peters2018deep}.
State-of-the-art LMs uses neural networks to calculate this probability \cite{bengio2003neural, mikolov2010recurrent, merity2016pointer, yang2017breaking}.

Within $X$, there exist a wide variety of words to be modeled, from closed-class function words, to common nouns or verbs, to named entities and numbers~\cite{zipf1949human}. %
Notably, words on the rarer end of this spectrum are often more semantically or topically important (as evidenced by the success of heuristics such as TF-IDF~\cite{salton1986introduction}, which up-weight words with low frequency).
Previous work has noted that while neural LMs greatly out-perform alternatives such as $n$-gram models on frequent words, they often under-perform on these rare words due to their limited parameter budget, which puts them at a disadvantage compared to non-parametric models like standard $n$-grams \cite{neubig2016generalizing}.
\begin{figure}[t]
\centering
\includegraphics[width=\linewidth]{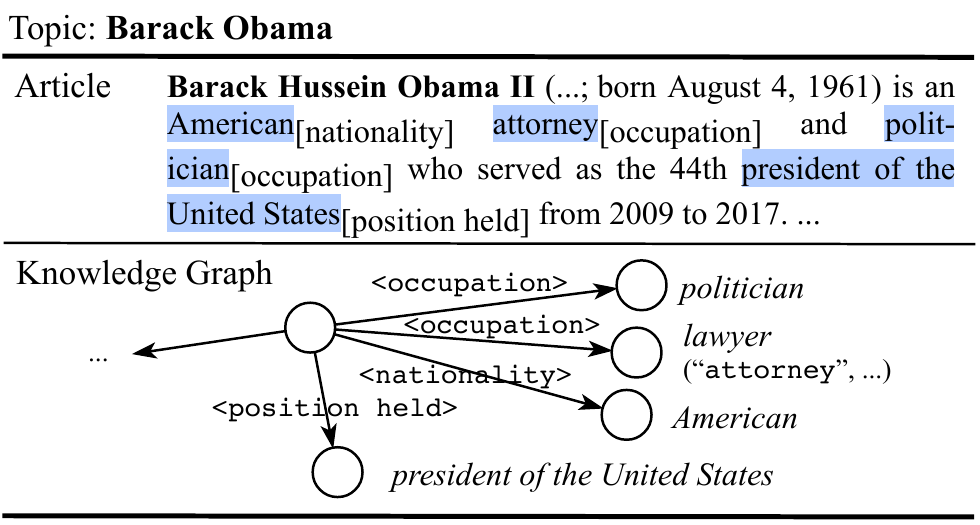}
\caption{Overview of our task of language modeling conditioned on structured knowledge. For a given topic, we want to learn an LM that leverages the
knowledge graph through relations when modeling the text.
}%
\label{fig:example}
\end{figure}
\footnotetext{\url{https://www.wikidata.org/wiki/Q892}.}

Ways to %
mitigate this bottleneck have been proposed in the context of \emph{conditional LMs}, which instead model the conditional probability $P(X \given C)$, where $C$ is some context given to the model. For instance in sequence transduction tasks,
there are mechanisms to copy from the source sequence~\cite{gu2016incorporating} or
use word or phrase dictionaries~\cite{arthur2016discretelexicons,tang2016neural} to
improve modeling of low-frequency words.
Perhaps more interesting from an LM perspective are methods explicitly conditioned on information from structured
knowledge sources such as knowledge graphs~\cite{angeli2010simple,Ahn2016ANK,parvez2018building,wang2018describing},
tables~\cite{barzilay2005collective, Lebret2016NeuralTG}, or grammars~\cite{konstas2013global}.
These methods are analogous to human language production, where the underlying knowledge or intent is converted into linguistic realizations.

In this work, we propose Latent Relation Language Models~(LRLMs), a class of conditional LMs
that take \textit{relational} information between entities in a knowledge graph as
context.
Specifically, our model is able to generate words either from a fixed word vocabulary, or through  spans defined according to their relations with a topic entity of interest, as shown in Figure~\ref{fig:example}.
The choice of which method of generation to use is defined as a latent variable sequence $Z$. We use Latent Predictor Networks~(LPNs; \newcite{ling2016latent}) to jointly learn $P(X, Z \given C)$, thus tractably marginalizing over all the possible spans.
Compared to other methods that condition LMs on knowledge graphs (KGs; \newcite{Ahn2016ANK,wang2018describing})
, the span-based generation from the KGs alleviates problems of malformed or incomplete mentions. Moreover,
the posterior probabilities of $Z$ can also be considered as entity links, which are of interest in
their own right in the information extraction field~\cite{ceccarelli2013learning, piccinno2014tagme, ganea2017deep}.

We apply the model on Wikipedia articles~($X$), with the help of relational information~($C$) such as Wikidata~\cite{vrandecic2014wikidata} or Freebase~\cite{bollacker2008freebase}
regarding each article topic.
Empirical results on open vocabulary language modeling show that the proposed model out-performs
previous approaches on the same task,
demonstrating that LRLMs provide an effective
way to condition on this context.
We also demonstrate the merit of explicitly modeling latent relations by examining the posterior
probabilities over the chosen relations $Z$, which are in concert with human intuitions about how
relations are being expressed in the text.

\section{Language Modeling Conditioned on Structured Knowledge}
\label{sec:taskndata}

First, we define the task of open-vocabulary language modeling conditioned on structured data.

\subsection{Task Definition}

Consider a directed and labeled knowledge graph~(KG) $G = \left( V, E\right)$
consisting of a set of nodes $V = \{v_1, \ldots, v_{|V|}\}$ and 
a set of relation edges $E = \left\{e_i\unpack\langle s_i, \omega_i, o_i \rangle \mid s_i, o_i \in V ,\ \omega_i \in R \right\}$.
Relation $e_i$ contains $s_i$, $\omega_i$, and $o_i$ as the subject, relation type, and object. $R$ is the set of all relation types.
Each node $v_i\in V$ represents either an entity or an attribute, and is associated with a set of surface forms
$\mathcal{A}(v_i) = \{a_{i,1},\ldots,a_{i,|\mathcal{A}(v_i)|}\}$ that can be used to refer to $v_i$.
For instance, the subject ``\textit{Barack Obama}''\footnote{\url{https://www.wikidata.org/wiki/Q76}.} is
connected to both ``\textit{politician}'' and ``\textit{lawyer}'' with the relation \texttt{<occupation>}, and the
object entity ``\textit{politician}''\footnote{\url{https://www.wikidata.org/wiki/Q82955}.} has ``\texttt{political figure}'' and
``\texttt{polit.}''\ as additional aliases. Notably surface forms of many objects in the KG can be multiple words, and thus it is necessary to have machinery to deal with this fact.

Given this KG, we further define a topic entity $s$ about which we would like to generate an explanation.
Our conditional language modeling problem is then defined as the problem of modeling the conditional probability of text $X$: $P(X \given G, s)$.
In particular, we consider 
a subgraph $G'=(V',E')$ of the original KG $G$ by extracting nodes and edges directly related
to the topic entity $s$:
\vspace{-1mm}
\begin{align*}
    V' &: \{s\} \cup \left\{o_i \mid \langle s, *, o_i  \rangle \in E \right\}, \\
    E' &:\{e_i\unpack\langle s, \omega_i, o_i\rangle \mid \langle s, \omega_i, o_i\rangle \in E \}.
\end{align*}

\begin{figure*}[t]
\centering
\includegraphics[width=\linewidth]{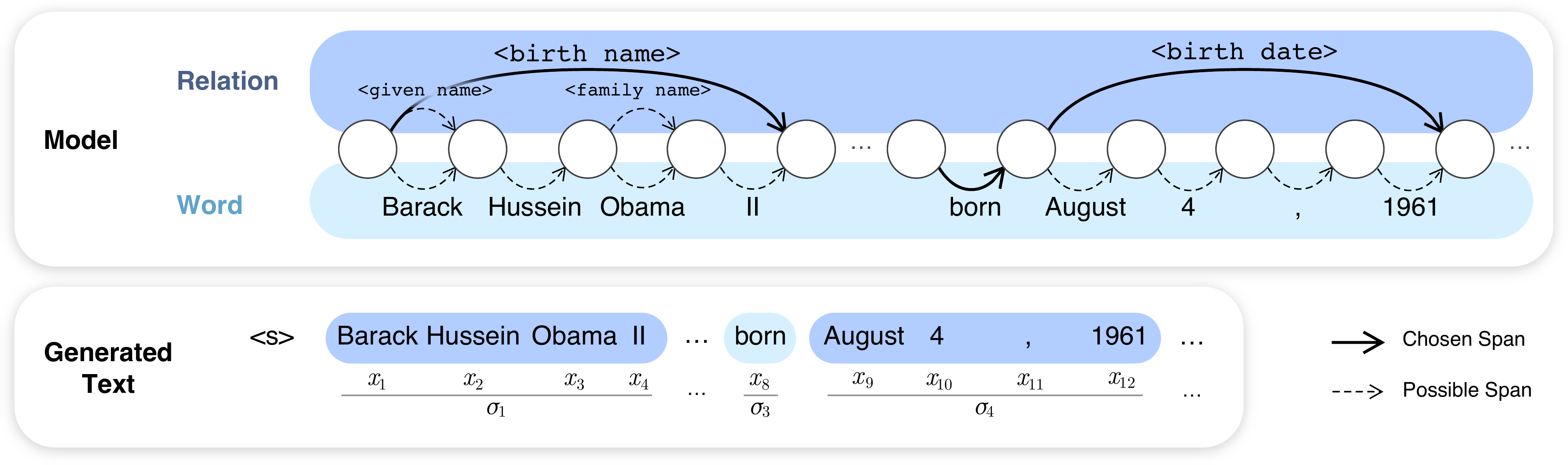}
\caption{While generating, our model switches between the two sources, namely ``Relation'' and ``Word''. Nodes represent hidden states up to each token, and edges represent possible span matches, \textit{i.e.}, choice of latent variables. In this example, we show one choice of latent variables with solid lines, and other options as dashed lines. We also show an ``annotation'' of the token sequence by the spans and sources we choose.}
\label{fig:architecture}
\end{figure*}

\vspace{-3mm}
\subsection{Why Condition on Knowledge Graphs?}
KGs provide two important benefits for neural LMs.
First, they have high coverage of rare words, which addresses lack of textual supervision for predicting these words. 
More importantly, KGs have the potential to help LMs generate \emph{factually consistent} text by providing 
factually consistent associations between entities.
Normal LMs would have to rely on supervision purely from textual data, which may not provide
a learning signal strong enough to accurately generate these facts.
For instance, results from \newcite{radford2019language} show that even with a very large model trained on massive amounts of data, samples
can be factually incorrect, although being fluent and coherent.

\section{Latent Relation Language Models}
\label{sec:model}

Next we describe our proposed framework of Latent Relation Language Models~(LRLMs).

\subsection{Motivation}
\label{sec:model:motivation}

The goal of the conditional language modeling task is to model the conditional probability $P(X\given G', s)$, assuming the presence of a KG subgraph $G'=(V',E')$ related to a topic entity~$s$.
Specifically, we can choose edges from $E'$ and copy the corresponding object nodes from $V'$. %
However, it is insufficient to model this probability using only $G'$ and $s$ as conditions, because it is unknown to us which text spans are matched to which relations, and simple text matching algorithms would yield many false positives.%
\footnote{For example, ``\textit{New York City}'' %
has an alias ``\texttt{New York}'', which matches ``\textit{New York}'' (state) and parts of ``\textit{New York City Council}''.}

To circumvent this lack of relation annotation, we treat such text spans as latent variables.
Formally, let $X = \{x_i\}_{i=1}^N$ be the sequence of $N$ tokens, and $Z = \{(\pi_t, \sigma_t, \rho_t)\}_{t=1}^T$ a sequence of latent variables describing text span matches:
\begin{itemize}[leftmargin=*, itemsep=0pt, topsep=6pt]
  \item The \textit{source} variable $\pi_t\in\{\textsc{rel}, \textsc{word}\}$ denotes the generation source of the span $x_{\sigma_t}$.
  \item The \textit{span} variable $\sigma_t=(\ell_t, r_t)$ specifies a token subsequence $x_{\sigma_t} = \{x_i\}_{i=\ell_t}^{r_t}$.
  \item The \textit{relation} variable $\rho_t=(e_t,a_t)$ describes the matching relation and surface form of the span $x_{\sigma_t}$, and is only used when $\pi_t=\textsc{rel}$.
\end{itemize}

For $Z$ to be a valid sequence of latent variables, the following conditions must be satisfied:
\begin{itemize}[leftmargin=*, itemsep=0pt, topsep=6pt]
  \item The span latent variables $\{\sigma_t\}_{t=1}^T$ form a \emph{segmentation} of $X$, \textit{i.e.}, $\ell_t=r_{t-1}+1$ for $t=2,\ldots,T$. This also implies $T\leq N$.
  \item If $\pi_t=\textsc{word}$, then $\ell_t=r_t$.
  \item If $\pi_t=\textsc{rel}$, then $\rho_t=(e_t,a_t)$ where $e_t=\langle s,\omega_t,o_t\rangle$ should satisfy $e_t\in E'$, $a_t\in \mathcal{A}(o_t)$, and $x_{\sigma_t}=a_t$, \textit{i.e.}, $\rho_t$ must correspond to a valid surface form of an object that is related to the topic entity $s$ and matches the text span.
\end{itemize}

Let $\mathcal{Z}$ be the set of all valid latent variable sequences. We can now model the conditional probability by marginalizing over $\mathcal{Z}$:
\begin{align}
  P(X \given G', s) & = \sum_{Z\in\mathcal{Z}} P(X, Z \given G', s). \label{eq:marginal}
\end{align}
We will show in section~\ref{sec:model:train} that this marginalization is tractable. For sake of brevity, unless noted otherwise, we drop $G'$ and $s$ from the conditions in the following sections.

\begin{algorithm*}[htbp]
\captionsetup{font=small,labelfont={small,bf}}
\caption{Generative Process of LRLM\label{alg:genstory}}
\algrenewcommand\algorithmicrequire{\bf Input}
\algrenewcommand\algorithmicensure{\bf Output}
\makeatletter
\newcommand{\linenum}{\arabic{ALG@line}}
\makeatother
\algrenewcommand\algorithmiccomment[1]{\hfill $\triangleright$ #1 {\small :\linenum}}
{\small
\begin{algorithmic}[1]
  \Require previous span $\sigma_{t-1}=(\ell_{t-1},r_{t-1})$, previously generated tokens $x_{<r_{t-1}}$.
  \Ensure source $\pi_t$, span $\sigma_t=(\ell_t,r_t)$, relation $\rho_t=(e_t,a_t)$, and token subsequence $x_{\sigma_t}$.
    \State $\ell_t \gets r_{t-1}+1$ \Comment{Update the beginning of span.} %
  \State $\widehat{\pi}_t \sim P(\pi_t \given x_{<\ell_t})$ \Comment{Choose whether to generate a word or relation.}
  \If { $\widehat{\pi}_t=\textsc{word}$} \Comment{Generating a word.}
    \State $P(\sigma_t, x_{\sigma_t},\rho_t\given \pi_t=\textsc{word},x_{<\ell_t}) \coloneqq P(x_{\ell_t}\given x_{<\ell_t})$ \Comment{Simplify the probability.}
    \State $\widehat{x}_{\ell_t} \sim P(x_{\ell_t}\given x_{<\ell_t})$ \Comment{Choose a word from model vocabulary.}
    \If { $\widehat{x}_{\ell_t} =$ \texttt{<UNK>}}
    \State $\widehat{x}_{\ell_t} \sim$ \textsc{CharModel} \Comment{Generate a word using a character model.}
    \ElsIf { $\widehat{x}_{\ell_t} =$ \texttt{<EOS>}}
       \State End generation.
    \EndIf
\ElsIf { $\widehat{\pi}_t=\textsc{rel}$} \Comment{Generating a relation.}
  \State $P(\sigma_t, x_{\sigma_t},\rho_t\given \pi_t=\textsc{rel},x_{<\ell_t}) \coloneqq P(e_t\given x_{<\ell_t})P(a_t\given e_t, x_{<\ell_t})$\label{eq:gen-rel} \Comment{Factor the probability.}
    \State $\widehat{e}_t \sim P(e_t\given x_{<\ell_t})$ \Comment{Choose a relation.}
    \State $\widehat{a}_t \sim P(a_t\given \widehat{e}_t, x_{<\ell_t})$ \Comment{Choose a surface form from the selected relation.}
    \State $\widehat{x}_{\sigma_t} \gets \widehat{a}_t$ \Comment{Generate a phrase.}
  \EndIf
\end{algorithmic}
}
\end{algorithm*}

\subsection{Definition}
\label{sec:model:def}

Given the latent variable sequence $Z$, we follow \newcite{ling2016latent} in factoring the joint probability:
\vspace{-5mm}
\begin{align*}
  P(X, Z)
  & = \prod_{t=1}^T P(\pi_t,\sigma_t,\rho_t,x_{\sigma_t}\given x_{<\ell_t}) \\
  & = \prod_{t=1}^T P(\pi_t\given x_{<\ell_t})P(\sigma_t,x_{\sigma_t},\rho_t\given \pi_t,x_{<\ell_t}),
\end{align*}
here $x_{<i}$ is the sequence of first $i-1$ tokens in $X$.
Figure~\ref{fig:architecture} shows an example of generation according to this factorization, and Algorithm~\ref{alg:genstory} precisely defines the process of generating at time step $t$.

\subsection{Training}
\label{sec:model:train}

During training, we marginalize over $\mathcal{Z}$ according to Equation~\ref{eq:marginal}.
Since the probability at time step $t$ is independent of previous latent variable choices, the marginalization is tractable using the forward-backward algorithm~\cite{baum1970maximization}.

Define the forward probability $\alpha_i$ as the marginal probability of the sequence up to the $i$-th token, computed as follows:
\begin{align*}
  \alpha_i = \sum_{(\pi,\sigma:(\ell,r),\rho)\in \tau_i}\alpha_{\ell} P(\pi,\sigma,x_{\sigma},\rho\given x_{<\ell}),
\end{align*}
where $\tau_i$ is the set of valid latent variable tuples $(\pi,\sigma\unpack(\ell,r),\rho)$ such that $r=i$, \textit{i.e.}, all valid spans ending at the $i$-th token. The marginal probability we optimize for is then $\alpha_N$. The backward probability $\beta_i$ which is required for gradient computation can be similarly calculated.

\vspace{-2mm}
\subsection{Parameterization}
\label{sec:model:param}

We use neural networks to parameterize all probability distributions mentioned above.
Decisions for time step $t$ are based on a $D$-dimensional hidden state $\textbf{h}_{\ell_t}$.
This hidden state can be generated by any neural sequence model, and we experiment with multiple models in experiments to demonstrate the generality of our approach.

\subsubsection{Source Selection}
Source selection is done using a simple linear model followed by a softmax function applied to the latest word-level hidden state $\mathbf{h}_{\ell_t}$:
\vspace{-1mm}
\begin{align*}
    P(\pi_t \given x_{<\ell_t}) = \mathrm{softmax}(\mathbf{W}_\pi \mathbf{h}_{\ell_t} + \mathbf{b}_\pi). %
\end{align*}
$\mathbf{W}_{\pi} \in \mathbb{R}^{2 \times D}, \mathbf{b}_{\pi} \in \mathbb{R}^{2}$ are trainable parameters.

\subsubsection{Word Generation}
\label{sec:model:wordgeneration}
Like conventional word-level neural language models, we have the option to generate the next token from a fixed vocabulary. This option is used to  generate any word that isn't an object participating in a relation.
The probability is:
\vspace{-1mm}
\begin{align*}
    P(x_{\ell_t} \given x_{<\ell_t}) &= \mathrm{softmax}(\mathrm{Linear}_w(\mathbf{h}_{\ell_t})), %
\end{align*}
where we define $\mathrm{Linear}(\mathbf{h})$ as a linear transform with a bottleneck of dimension $K$ into a vector over vocabulary size $L$:
\begin{align}
    \mathrm{Linear}(\mathbf{h}) = \mathbf{W}_1(\mathbf{W}_2\mathbf{h} + \mathbf{b}_2) + \mathbf{b}_1, \nonumber
\end{align}
where $\mathbf{W}_1 \in \mathbb{R}^{L \times K}$, $\mathbf{b}_1 \in \mathbb{R}^{L}$, $\mathbf{W}_2 \in \mathbb{R}^{K \times D}$, $\mathbf{b}_2 \in \mathbb{R}^{D}$ are trainable parameters.
Empirically we found this low-rank version to out-perform a full linear transform.

\paragraph{Generating unknown words}
As our task is open-vocabulary language modeling, we must be able to generate words even if they are out of vocabulary.
Following \newcite{chung2016hierarchical} and \newcite{luong2016achieving},
we do so by having a character-level LM ``spell-out'' any unknown words.
If the unknown word is $x = c_1\ldots c_{|c|}$ with $|c|$ characters: %
\begin{align}
    P(x \given x_{<\ell_t}) = P(\texttt{<UNK>} \given x_{<\ell_t}) P(c_1\ldots c_{|c|} ; \theta_{\text{char}}),\nonumber 
\end{align}
where $\theta_{\text{char}}$ are the parameters of the character LM. We pre-train this model on the
set of all the unique words in the training set and fix its parameters while training LRLM.

\subsubsection{Relation Generation}
The goal of relation generation is to find the most suitable span that can be copied 
into the text. As Line~\ref{eq:gen-rel} of Algorithm~\ref{alg:genstory} depicts, this probability is  factorized into two steps: relation selection and surface form selection.

\paragraph{Relation selection}
We utilize pretrained KG embeddings\footnote{Specifically, from OpenKE~\cite{han2018openke}.} for entities and relation types. For a relation $e_i\unpack\langle s, \omega_i, o_i \rangle$, we concatenate KG embeddings for $\omega_i$ and $o_i$ to obtain the \textit{relation embedding} $\mathbf{e}_i$.%
\footnote{We train embeddings for each relation type not covered by pre-trained embeddings, and an UNK embedding for attributes and entities not covered by pre-trained embeddings.}
We then compute the probability of selecting each relation as:
\begin{align*}
  P(e_i\given x_{<\ell_t}) = \mathrm{softmax}(\mathbf{e}_i^\top\mathrm{Linear}_o(\mathbf{h}_{\ell_t})). %
\end{align*}

\paragraph{Surface form selection}
We featurize surface forms via fastText~\cite{bojanowski2017enriching} embeddings pre-trained on
the training corpus, and calculate probability of surface form $a_k$ as:
\begin{align*}
    P(a_k \given e_i, x_{<\ell_t}) = \mathrm{softmax}(\mathbf{f}_{a_k}^\top(\mathbf{W}_a \mathbf{h}_{\ell_t} + \mathbf{b}_a)),
\end{align*}
where $\mathbf{f}_{a_k}$ is the embedding for $a_k$ and $\mathbf{W}_a$, $\mathbf{b}_a$ are trainable parameters.

\begin{table}[t]
\centering
\resizebox{\columnwidth}{!}{%
\begin{tabular}{lcrrrr}
\toprule
Dataset           & Doc & Vocab & Rel/Ent & Tok/Doc  & Ment/Doc  \\ \midrule
WikiFacts  & \phantom{0}7856  & 40.0k & 82.71 & 157.25  & 9.64        \\
WikiText-S & 27685 & 71.1k & 11.38 & 295.75  & 11.20       \\
WikiText-F & 27685 & \phantom{.}264k  & 11.38 & 3559.91 & 73.01       \\\bottomrule
\end{tabular}
}
\caption{Training set statistics for all dataset variations: number of training documents, vocabulary size, relations per head entity, tokens per document, and entity mentions per document.
\vspace{-3mm}
}

\label{tab:datastats}
\end{table}

\section{Datasets}
\label{sec:data}

We use two datasets with different characteristics for experiments; statistics are shown in Table~\ref{tab:datastats}.

\subsection{WikiFacts}
WikiFacts\footnote{{\small \url{https://bitbucket.org/skaasj/wikifact_filmactor}}}%
~\cite{Ahn2016ANK} is a collection of Wikipedia articles restricted to 
\texttt{/film/actor} domain entities in Freebase~\cite{bollacker2008freebase}. %
Each example consists of the \textit{first section} of the original article.
Since official splits for evaluation are not provided, we follow previous work and performed a random split of 
80/10/10\%.%

In addition to Freebase, this dataset expands the set of relations by including topic
entities from other articles linked to the page to be generated.
Since these (gold) entities will not be available if we attempt to generate new articles,
we remove them from the dataset for our main experiments\footnote{For consistency with prior work, we also report results with them in Appendix~\ref{app:anchor}.}.

Finally, we note that this dataset does not include aliases for entities, \textit{i.e.}, $|\mathcal{A}(o)|=1$ for all objects $o$.
Hence, the surface form selection module acts as oracle, where it always assigns a probability of 1 to the correct surface form.

\subsection{WikiText}
While WikiFacts has been used in previous work on LMs using structured data~\cite{Ahn2016ANK}, the domain is limited (film actors). To investigate the capability of knowledge-infused LMs in an open-domain setting with a wide variety of relations, we build a large-scale open-domain
dataset from the existing WikiText-103 dataset~\cite{merity2016pointer} by associating articles with entities in Wikidata~\cite{vrandecic2014wikidata}.
We employ the same data splits from the original dataset.
In the following paragraphs, we discuss how we bridge KGs and the articles from WikiText-103 (more details in Appendix~\ref{app:datacollection}).
\paragraph{Constructing subgraphs for articles}
As discussed in Section~\ref{sec:taskndata}, we take the original KG and extract a relevant subgraph $G'$
for each article. While there are many options on how to extract this subgraph, we choose the subgraph $G'$ consisting of \textit{direct neighbors} of the topic entity for each article.
This forms a star-shaped subgraph, with the topic entity as the central node, connected by the
related entities and attributes. 
We found on average 3.1 surface forms for each entity.

\paragraph{Linking mentions with the KG}
For each object in $G'$, we search for occurrences of all surface forms in the article while allowing
token overlaps among them. %
Note that, similarly to distant supervision for relation extraction \cite{mintz2019distant}, this process can produce false positive relation mentions because of simple string-based matching.
We rely on our model's ability to ignore such mentions by learning
to assign high probabilities only on the correct mentions.

We name the dataset obtained through this process WikiText-F (Full). We also create WikiText-S
(Short) by truncating after the first sections of each example in WikiText-F.
This dataset is similar to WikiFacts in terms of article length, and allows
performance comparisons among the two datasets.

\section{Experiments}
\label{sec:experiment}
As previously noted, we evaluate our models on \textit{open-vocabulary language modeling} and report token-level perplexity. This provides more realistic perplexity measures of text than in closed setting by considering OOV words.
Specifically, we use pre-trained character-level LMs from Section~\ref{sec:model:wordgeneration} for each dataset to discount the probability of an unknown word based on its spelling. Unlike UPP~\cite{ueberla1994analysing}, which also adjusts the perplexity of OOV words but are limited within corpus, discounting based on spelling enables truly open-vocabulary evaluation.
This is done for all tested models, both proposed and baselines.

\subsection{Model Configuration}
For WikiFacts, we use a fixed word vocabulary size of 40,000 following previous work.
For WikiText-derived datasets, we include all words with frequencies no less than 3 in our dataset
following~\citet{merity2016pointer}. We use adaptive embeddings and softmax to handle large vocabulary~\cite{baevski2018adaptive,grave2016efficient}.%

To calculate the hidden state $\mathbf{h}_{x_{<i}}$, we test two varieties of neural sequence models: standard LSTMs~\cite{hochreiter1997long}, and the state-of-the-art Transformer-XL~\cite{dai2019transformer}.
We implement all models in PyTorch~\cite{paszke2017automatic}.
Training details and hyperparameters are summarized in Appendix~\ref{app:hyperparams}.

\begin{table*}
\centering
{
\small
\begin{tabular}{llcclp{0pt}ccl}
\toprule
\multirow{2}*{Base model}     & \multirow{2}*{Dataset} & \multicolumn{3}{c}{Dev} & & \multicolumn{3}{c}{Test} \\ \cmidrule{3-5}\cmidrule{7-9}
                              &           & \multicolumn{1}{c}{Vanilla LM} & NKLM   & LRLM   & & \multicolumn{1}{c}{Vanilla LM} & NKLM    & LRLM  \\ \midrule
\multirow{3}*{LSTM}           & WikiFacts       & 219.11 & 93.09   & \bf 89.55$^{\ast}$  & & 208.44 & 87.88   & \bf 82.89$^\ast$  \\ 
                              & WikiText-S      & \phantom{0}68.37  & 46.16   & \bf 45.84           & & \phantom{0}86.12  & 55.98  &  \bf 55.38    \\
                              & WikiText-F      & \phantom{0}45.13  & 44.46   & \bf 42.18$^{\ast}$  & & \phantom{0}49.47  & 48.54   & \bf 45.70$^{\ast}$  \\ \midrule
\multirow{3}*{Transformer-XL} & WikiFacts       & 170.40 & 98.98   & \bf 83.19$^{\ast\ast}$  & & 162.65 & 92.92   & \bf 76.46$^{\ast\ast}$  \\
                              & WikiText-S      & \phantom{0}42.63  & 43.05   & \bf 37.75$^{\ast\ast}$  & & \phantom{0}52.96  & 52.51   & \bf 44.98$^{\ast\ast}$  \\
                              & WikiText-F      & \phantom{0}30.14  & 32.19   & \bf 29.56$^{\ast\ast}$  & & \phantom{0}33.01  & 35.27   & \bf 32.20$^{\ast\ast}$  \\ \bottomrule
\end{tabular}
}
\caption{Perplexity values of different models on open vocabulary language modeling, lower is better. Best results are in bold. Asterisk symbols represent statistical
    significance according to Wilcoxon signed-rank test~\cite{dror2018hitchhiker} against the better model among \textsc{NKLM} and Vanilla LM, with $p < 0.05$ ($^\ast$) and $p < 0.01$ ($^{\ast\ast}$), respectively.}
\label{tab:ppl}
\end{table*}

\subsection{Baselines}
We compare LRLM against two baselines:

\paragraph{Vanilla language model (Vanilla LM)} %
This is a simplification of LRLM removing the relation generation module, analogous to standard LSTM or Transformer-XL language models from previous work~\cite{merity2017regularizing,dai2019transformer}.

\paragraph{Neural Knowledge Language Model (NKLM)}
Similar to LRLM, the Neural Knowledge Language Model (NKLM; \newcite{Ahn2016ANK}) also has the ability to copy from a given set of KG triples, but differs from LRLM in several ways:
\begin{enumerate}[leftmargin=*, itemsep=1pt, parsep=2pt, topsep=6pt]
    \item LRLM marginalizes over all derivations of a sequence, which allows processing of overlapped tokens among spans, while NKLM makes all decisions in a hard fashion and cannot handle such overlapped tokens.\footnote{We perform additional data preprocessing on WikiText for NKLM, detailed in Appendix~\ref{app:nklm-preprocess}.}
    \item LRLM allows generation at span-level (\textit{i.e.} can predict multi-word entities at once), while NKLM predicts one word at a time and the model needs to repeatedly predict the right relation until copying of an object is done.
\end{enumerate}

The original NKLM does not differentiate between aliases, so we perform the same surface form selection as LRLM for fair comparison.

\begin{figure*}[t]
\centering
\includegraphics[width=\linewidth]{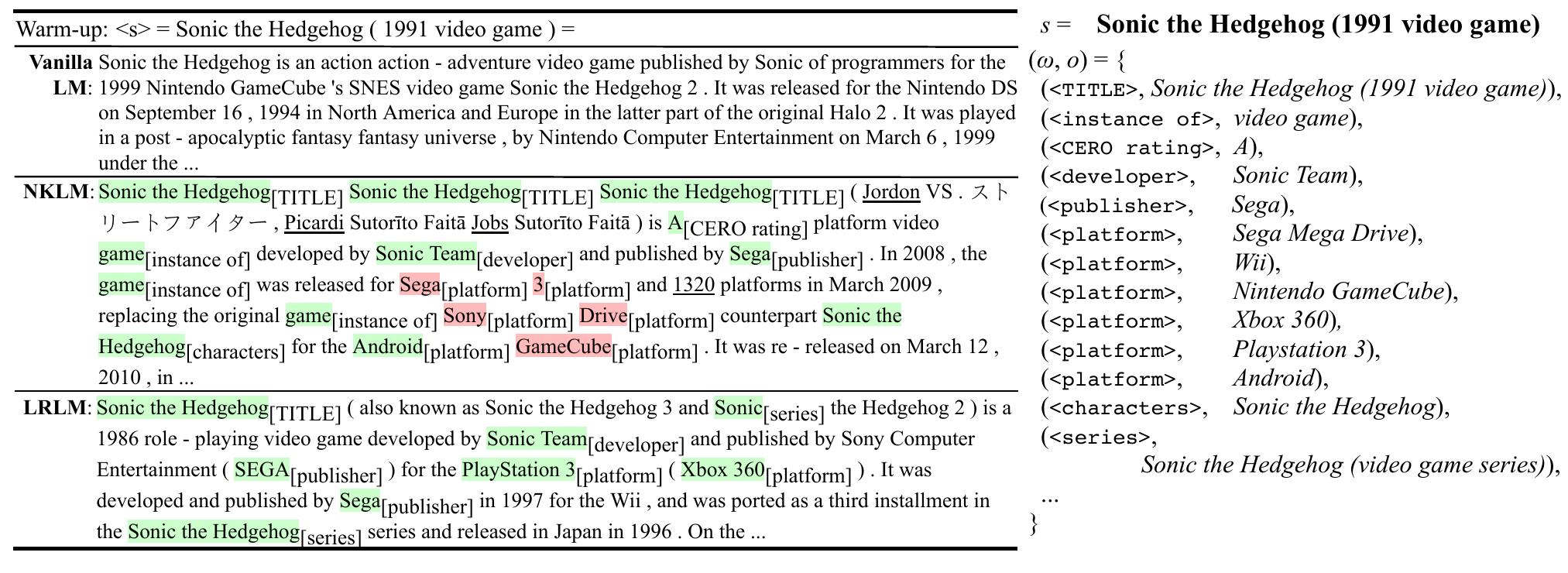}
\vspace{-7mm}
\caption{Samples from the three models for a topic entity ``\textit{Sonic the Hedgehog (1991 video game)}'' with the corresponding subgraph on the right.
  Square brackets denote the relation type of copied objects.
  Highlighted spans in light green represent objects that are copied in full, whereas those
  in dark red represent partially copied objects. Underlined tokens are unknown words sampled
  from character model.} %
\label{fig:sample}
\end{figure*}

\section{Results and Analysis}
\label{sec:analysis}

\subsection{Main Results}
\label{sec:results}
Perplexities over the datasets are shown in Table~\ref{tab:ppl}.
We observe that for both sequence models, LRLM out-performs the baselines on all datasets (although on the one case of LSTM+WikiText-S the improvement was not statistically significant).
Particularly on the two WikiText-derived datasets, our model shows significant improvements
over the baselines by leveraging KGs in comparison to the vanilla LM, while NKLM has difficulty
utilizing the KGs to achieve better perplexity, and in some cases results in worse perplexities than the vanilla LM.
Note that these results are on open-vocabulary modeling, and results and analyses on the closed vocabulary setting can be found in Appendix~\ref{app:anchor}. We also report UPP values~\cite{ueberla1994analysing} in Appendix~\ref{app:upp}.

\subsection{Generated Samples}
To illustrate the behavior of the learned models, we take the three models trained on WikiText-S %
and draw 10 samples while conditioning on $G'$ and $s=\text{``\textit{Sonic the Hedgehog}''}$, and show the sample with lowest perplexity in Figure~\ref{fig:sample}.
Highlighted terms with different colors represent two types of mentions generated from the relation predictor: full and partial.
A \textit{full mention} is an identical copy of an entity surface form, while a \textit{partial mention} is an
incomplete subphrase of an entity surface form. NKLM's word-by-word generation scheme results in
partial mention being generated, while LRLM does not due to span-level copying from KGs.
A perfect model should not generate partial mentions as it leads to possibly corrupted phrases,
and should generate the same set of full mentions as the gold mentions.

Although NKLM generates more mentions, it suffers from generating partial mentions because it 1) is unaware of the length of entities, and 2) requires making copy decisions as many times as the number of tokens in a phrase.
As a result, we often observe NKLM switching entities or surface forms halfway through, ending mentions early, and repeating the same entity.
In contrast, LRLM, by design, only generates full mentions.

We quantitatively show this in Table~\ref{tab:copystats} by counting the average number of partial and full mentions 
in samples. We take 10 samples from 10 random development set articles.
Next, we performed a precursory manual annotation of ``valid'' mentions, which we deemed as semantically correct based on the sentential context.
NKLM generates more invalid mentions than LRLM, most of which are false positives and repetitions of the same entity.
LRLM has almost no repetitions, but sometimes incorrectly predicts the article's theme.\footnote{For example, generating an article about a TV episode for a topic entity of a song.}

\begin{table}[t]
\centering
{
\small
\begin{tabular}{r|rr|rr}
\toprule
     & Partial & Full & Valid & Invalid \\ \midrule
NKLM &    16.9 & 7.81 &  6.37 & 1.44  \\
LRLM &    $-$  & 6.32 &  5.63 & 0.69 \\ 
Gold &    $-$  & 9.00 &  9.00 & 0.00\\ \bottomrule
\end{tabular}
}
\caption{\label{tab:copystats} Average number of partially generated, fully generated, and valid mentions over 100 samples from the development set or gold human-generated article.}
\vspace{-2mm}
\end{table}

\subsection{Posterior Probability of Spans}
\label{sec:analysis:posterior}
One of the advantages of our model is its capability to calculate the posterior probability of a relation
generating a span in an existing text. We calculate the joint probability of a span and the surrounding
text\footnote{We consider the text segment in the batch where the span belongs to as the surrounding text.}
by marginalizing over the latent variable $Z$ for both sides of context, and normalize over all possible spans:
\begin{align*}
    P(X, Z) &= \alpha_i \cdot P(Z \given x_{<\ell_{i}}) \cdot \beta_i\\
    P(Z \given X) &= P(X, Z) \,/\,\sum_{Z \in \mathcal{Z}} P(X, Z)
\end{align*}
where $\beta_i$ is the backward probability calculated reversely following Section~\ref{sec:model:train}.
Table~\ref{tab:posterior} shows spans with the posterior probability of various relation types from an article about ``\textit{Sorry (Madonna song)}''. %
The model demonstrates the ability to relate the entity ``\textit{Madonna}'' to the topic based on context.
We also observe a general trend that the model prefers generating multi-word spans through relations rather than word by word from vocabulary.
However, when generating common phrases (\textit{e.g.}, ``\textit{the United States}''), our model often favors word-based generation even if an alternative
relation-based prediction is possible.

\begin{table}[t]
\centering
\small
\setlength{\tabcolsep}{1pt}
\begin{tabular}{rrrrr}
\toprule
\multicolumn{5}{l}{Title: Sorry (Madonna Song)}\\\midrule
\multicolumn{5}{c}{... song by American singer \colorbox{green!20}{Madonna} from her tenth ... } \\
\multirow{3}*{Relations:}& \multicolumn{2}{r}{\texttt{<performer>}} & \multicolumn{2}{c}{\textbf{0.9697}}   \\
                        & \multicolumn{2}{r}{\texttt{<lyrics by>}} & \multicolumn{2}{c}{0.0289}          \\
                        & \multicolumn{2}{r}{\texttt{word}}        & \multicolumn{2}{c}{0.0014}           \\ \midrule
 \multicolumn{5}{c}{... written and produced by \colorbox{green!20}{Madonna} and Stuart Price , ...} \\
\multirow{3}*{Relations:} & \multicolumn{2}{r}{\texttt{<performer>}} & \multicolumn{2}{c}{0.1545}           \\
                         & \multicolumn{2}{r}{\texttt{<lyrics by>}} & \multicolumn{2}{c}{\textbf{0.7693}}   \\
                         & \multicolumn{2}{r}{\texttt{word}}        & \multicolumn{2}{c}{0.0762}           \\ \midrule
 \multicolumn{5}{c}{... continuation from the `` \colorbox{green!20}{Hung Up} '' music video . ...} \\
\multirow{2}*{Relations:} & \multicolumn{2}{r}{\texttt{<follows>}} & \multicolumn{2}{c}{\textbf{1.0000}}           \\
                         & \multicolumn{2}{r}{\texttt{word}}      & \multicolumn{2}{c}{0.0000}           \\ \midrule
 \multicolumn{5}{c}{... . However , in \colorbox{green!20}{the United States} , the song did ...} \\
\multirow{3}*{Relations:} & \multicolumn{2}{r}{\texttt{<origin>}}  &  \multicolumn{2}{c}{0.0000}   \\
                          & \multicolumn{2}{r}{\hspace{1em}\texttt{word} $\rightarrow$ \texttt{<origin>}} &  \multicolumn{2}{c}{0.0003} \\
                         & \multicolumn{2}{r}{\texttt{word}}         & \multicolumn{2}{c}{\textbf{0.9997}}          \\ \bottomrule
\end{tabular}
\vspace{-2mm}
\caption{\label{tab:posterior}Posterior probability of spans (highlighted) in contexts. \texttt{word} represents
    word-based generation. The second relation in the last example means generation of ``\textit{the}''
    using \texttt{word}, followed by relation-based generation of ``\textit{United States}'' using the
    \texttt{<origin>} relation.%
\vspace{-1mm}
}
\end{table}

\subsection{Effect of Subgraph Size}
\begin{figure}
  \centering
  \includegraphics[width=0.96\linewidth, keepaspectratio]{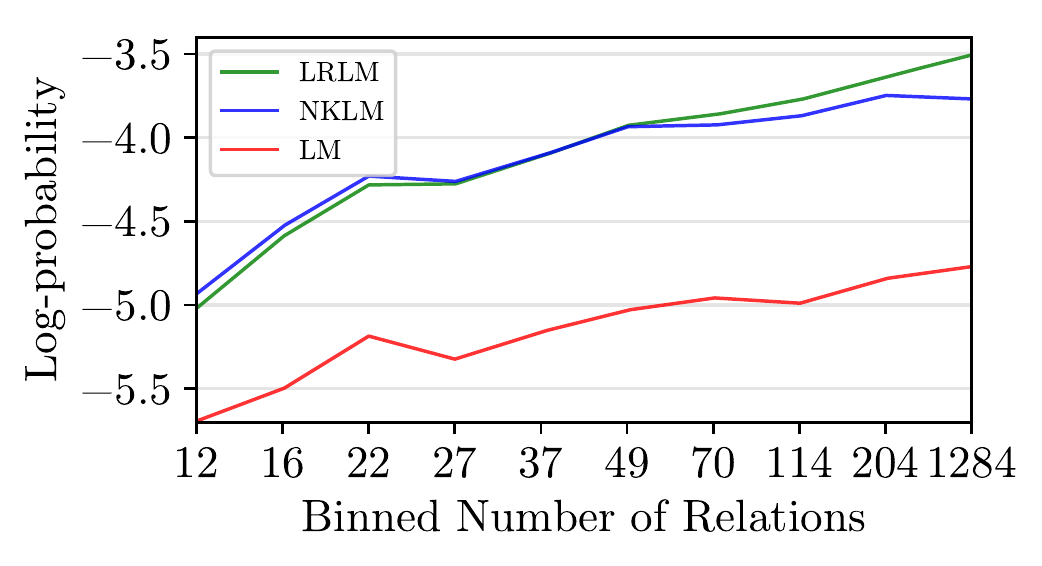}
  \vspace{-3mm}
  \caption{\label{fig:subgraphsize}Word-average log-probabilities on development set of WikiFacts grouped by average relations per article. LRLM shows a larger gain over the baselines as the number of relations increases.}%
  \vspace{-2mm}
\end{figure}

Finally, we measure the performance of models with respect to the richness of resource available for conditioning.
We group WikiFacts articles into 10 bins by the number of relations available, and plot binned word-average log-probabilities %
in Figure~\ref{fig:subgraphsize}.
While all models have slightly higher log-probabilities as the number of relations increase,
LRLM achieves the largest gain. %
We believe this is due to marginalization over the latent variables in LRLM helping better disambiguate between many candidates, while NKLM struggles to predict the right relations and surface form lengths as the number of candidates increases.

\vspace{-2mm}
\section{Related Work}
\label{sec:relwork}

\vspace{-2mm}

A variety of entity-aware LMs exist, conditioning on a variety of information sources such as expert coreference annotations~\cite{ji2017dynamic, clark2018neural,yang2017reference}, entity annotations~\cite{logan2019baracks}, definitions
\cite{bahdanau2017learning}, or keywords~\cite{kiddon2016globally, parvez2018building}.
As mentioned above, NKLM~\cite{Ahn2016ANK} is the most relevant previous work that uses relational information.
Our proposed LRLM formulation is more successful at lowering perplexity and also allows calculating posterior probabilities of relations.

Incorporating KGs for natural language generation~(NLG) has a long history~\cite{goldberg1994using,reiter2005choosing,chen2008learning}. %
With the recent advancement of neural sequence modeling, prevalent approaches for
language generation from KGs employ sequence-to-sequence models~\cite{sutskever2014sequence}
with special attention mechanisms tailored for input structures such as 
graphs~\cite{wang2018describing} or tables~\cite{liu2017table,perez2018bootstrapping}.
Unlike our focus, however, this class of research focuses on learning discriminative models that do
not explicitly generate the referent entity as latent variables, like we do in Section~\ref{sec:analysis:posterior}.

While not directly related to our core task, there have been a number of other methods for incorporating latent variables into NLG problems.
Latent structure has included predicting latent sequences of topics~\cite{wiseman2018learning},
chunking of word sequences into $n$-grams~\cite{Buckman2018NeuralLL}, deciding between input
sources~\cite{ling2016latent, gu2016incorporating}, predicting latent continuous vectors~\cite{bowman2016continuousspace},
generating compressed summary tokens~\cite{miao2016languageaslatent},
or inducing syntactic and semantic trees~\cite{yogatama2016learning,yin2018structvae}.
Our work borrows heavily from \citet{ling2016latent}, who select from multiple sources for source code generation.
We use a similar method for selecting latent sources for Wikipedia
article language modeling with a repository of KG triples.
\vspace{-2mm}

\section{Conclusion}
\label{sec:conclusion}
In this work, we propose Latent Relation Language Models, a class of conditional language models
conditioned on knowledge graphs. Our generative framework models text as a sequence of spans, some
of which are generated as entities included in the knowledge graph. Marginalization over latent variables
allows the model to not only out-perform previous work in conditional language
modeling tasks, but also score spans with their posterior relation probability.

\subsection*{Acknowledgements}
This research was supported in part by Funai Foundation for Information Technology and Amazon.
The authors would also like to thank Qian Wang for helping designing the model figure and the members of the NeuLab for helpful discussion. 

\bibliography{emnlp-ijcnlp-2019}
\bibliographystyle{acl_natbib}

\clearpage
\appendix

\section{Article Collection}
\label{app:datacollection}
We collect seed Wikipedia articles from the raw release of WikiText-103~\cite{merity2016pointer},
where raw vocabulary was preserved.
Minimal preprocessing was performed by the dataset providers.\footnote{Data can be found at \url{https://s3.amazonaws.com/research.metamind.io/wikitext/wikitext-103-raw-v1.zip}.}
The dataset provides
an open domain, quality-assured set of Wikipedia articles verified by editors. We take the dataset and split
each set back into per-article texts with simple regular expression rules for detecting titles.
Then we query the Wikipedia API to identify the Wikidata entity\footnote{We used a Wikidata dump as of 2018/09/20.}
for each article.
During this process, we discarded some articles in the training set where the API failed to return
Wikidata IDs, which was due to their deletion or title renames since the release of original dataset in 2016.
For development and test set, we manually matched the few missed articles to recover all the articles.

\begin{table}[t]
\centering
\small
\begin{tabular}{rr}
\toprule
\multicolumn{2}{c}{Common hyperparameters} \\\cmidrule{1-2}
Learning rate decay rate & 0.9 \\
Batch size & 60 \\
BPTT window size & 150 \\
Entity embedding size & 50/100/100 \\
fastText embedding size & 300 \\ \midrule
\multicolumn{2}{c}{Transformer-XL hyperparameters} \\ \cmidrule{1-2}
Learning rate & 0.00025 \\
Warm up steps & 6000 \\
Attention dropout rate & 0 \\
Dropout rate & 0.1  \\
Embedding size & 410  \\
FC layer hidden unit size & 2100 \\
Memory size & 150 \\
Number of layers & 16 \\
Number of heads & 10 \\
Per-head attention dimension & 41 \\ \midrule
\multicolumn{2}{c}{LSTM hyperparameters} \\ \cmidrule{1-2}
Learning rate & 0.001 \\
Dropout rate & 0.5 / 0.5 / 0.1  \\
Embedding size & 400 / 400 / 512  \\
Hidden unit size & 1000 / 1000 / 1024  \\
Linear hidden unit size & 1000 / 1000 / 500  \\
Number of layers & 2/2/4 \\ \midrule
\multicolumn{2}{c}{LRLM-specific hyperparameters} \\ \cmidrule{1-2}
Relation linear hidden unit size & 1000 / 1000 / 800  \\ \midrule
\multicolumn{2}{c}{NKLM-specific hyperparameters} \\ \cmidrule{1-2}
Max position count & 20 \\
Position embedding size & 40 / 40 / 50 \\ \bottomrule
\end{tabular}
\caption{Model and training hyperparameters that are common across the models. Slash-delimited values
represent different hyperparameters used in WikiFacts, WikiText-S, WikiText-F, respectively.}
\label{tab:app:hyperparams}
\end{table}

\section{Training Details and Hyperparameters}
\label{app:hyperparams}
\paragraph{Training Details}
All models are trained using Adam~\cite{kingma2014adam}.
Models equipped with Transformer-XL are trained with the same schedule as the original paper; the learning rate is linearly increased over the first 6000 gradient steps up to 0.00025, and reduced
according to cosine annealing. Models with LSTM are trained with the initial learning rate set to 0.001. Validation is performed on the development set after every epoch, and when validation loss does not improve, learning rate is multiplied by 0.9 and the model and optimizer parameters are reset to the previous checkpoint.
For all experiments, we use truncated backpropagation through time~\cite{williams1990efficient} with the truncation window size being 150.
\paragraph{Hyperparameters} 
We list the common model hyperparameters for the model in Table~\ref{tab:app:hyperparams}.
While we use the same Transformer-XL hyperparameters across datasets, we apply different sets of LSTM
hyperparameters on WikiFacts, WikiText-S, and WikiText-F for better performance. %
See Section~\ref{sec:experiment} for more details on the vocabulary size.
We take pre-trained KG embeddings from OpenKE~\cite{han2018openke}, with dimensions of 50 and 100 for WikiFacts and WikiText respectively.\footnote{\citet{Ahn2016ANK} uses 100-d KG embeddings, but there were no publicly available embeddings in that dimension.}

\section{Utilization of Extra Entities}
\label{app:anchor}

\begin{table*}
\centering
\small
\begin{tabular}{lcccp{0pt}ccc}
\toprule
\multirow{2}*{Dataset} & \multicolumn{3}{c}{Dev} & & \multicolumn{3}{c}{Test} \\ \cmidrule{2-4}\cmidrule{6-8}
  & \multicolumn{1}{c}{Vanilla LM} & NKLM    & LRLM   & & \multicolumn{1}{c}{Vanilla LM} & NKLM    & LRLM  \\ \midrule
WikiFacts
  & 217.19 & 95.68   & \bf 94.64  & & 207.54 & 90.44   & \bf 87.73  \\ \midrule
$+$ Entity
  & 217.19 & 59.84   & \bf 54.60  & & 207.54 & 57.14   & \bf 51.34  \\
$+$ Oracle char model
  & \phantom{0}88.03  & 38.54   & \bf 34.73  & & \phantom{0}84.56  & 37.23   & \bf 33.02  \\ \midrule
\cite{Ahn2016ANK}
  & \phantom{0}82.4\phantom{0}   & 41.4\phantom{0}    & --  & & \phantom{0}86.4\phantom{0}   & 43.6\phantom{0}    & --  \\ \bottomrule
\end{tabular}
\caption{Perplexity values of models on WikiFacts, lower is better. ``$+$ Entity'' means trained with extra entities; ``$+$ Oracle char model'' means treating the character model as oracle, \textit{i.e.}, treating spell-out probabilities of OOV words as 1. Best results are in bold. Note that our results are not directly comparable with reported results by \citet{Ahn2016ANK} due to different dataset splits being used.}
\label{tab:anchor}
\end{table*}

\begin{table*}[htbp]
\centering
{
\small
\begin{tabular}{llcclp{0pt}ccl}
\toprule
\multirow{2}*{Base model}     & \multirow{2}*{Dataset} & \multicolumn{3}{c}{Dev} & & \multicolumn{3}{c}{Test} \\ \cmidrule{3-5}\cmidrule{7-9}
                              &           & \multicolumn{1}{c}{Vanilla LM} & NKLM   & LRLM   & & \multicolumn{1}{c}{Vanilla LM} & NKLM    & LRLM  \\ \midrule
                              \multirow{3}*{LSTM}           & WikiFacts       & 156.29 & 74.04   & \bf 71.20$^{\ast}$  & & 148.05  & 70.08   & \bf 66.09$^{\ast}$ \\ 
                              & WikiText-S      & \phantom{0}65.42  & 49.95   & \bf 44.44$^{\ast\ast}$          & & \phantom{0}80.69  & 60.96  &  \bf 52.81$^{\ast\ast}$ \\
                              & WikiText-F      & \phantom{0}43.59  & 42.99   & \bf 40.88$^{\ast\ast}$  & & \phantom{0}47.14  & 46.37   & \bf 43.72$^{\ast\ast}$  \\ \midrule
\multirow{3}*{Transformer-XL} & WikiFacts       & 121.55 & 78.72   & \bf 66.14$^{\ast\ast}$  & & 115.53 & 74.09   & \bf 60.96$^{\ast\ast}$  \\
                              & WikiText-S      & \phantom{0}40.79  & 41.59   & \bf 37.75$^{\ast\ast}$  & & \phantom{0}49.62  & 49.92   & \bf 42.76$^{\ast\ast}$  \\
                              & WikiText-F      & \phantom{0}29.11  & 32.19   & \bf 28.59$^{\ast\ast}$  & & \phantom{0}31.45  & 33.69   & \bf 30.75$^{\ast\ast}$  \\ \bottomrule
\end{tabular}
}
\caption{UPP of different models, lower is better. Best results are in bold. Asterisk symbols represent statistical
    significance according to Wilcoxon signed-rank test~\cite{dror2018hitchhiker} against the better model among \textsc{NKLM} and Vanilla LM, with $p < 0.05$ ($^\ast$) and $p < 0.01$ ($^{\ast\ast}$), respectively.}
\label{tab:pplupp}
\end{table*}

Adding extra entities to WikiFacts increased the average number of relations per article from 82.71 to 89.28, and mentions from 9.64 to 16.97. On average, each added entity matches 1.12 spans.

Table~\ref{tab:anchor} compares results under different settings.
The inclusion of extra entities significantly improves results for both models. This is due to the fact
that these entities are extracted from hyperlinks within text, so 1) they are mostly rare words; 2) the model
can easily learn that all such entities must be included in the text at some point.

\section{Data Preprocessing for NKLM}
\label{app:nklm-preprocess}

\paragraph{WikiFacts}
The provided WikiFacts dataset contains KG subgraphs and text annotated with non-overlapping matched spans. Copying positions%
\footnote{Copying position of a word is the 0-based word index into the matching entity surface form, which indicates the position to copy from.}
are sequentially assigned within a matched span.

One caveat is that the dataset includes relations containing Freebase \textit{Compound Value Type}~(CVT) as entities.
These types are used to encapsulate a structured representation with multiple fields. We removed all relations where the subject entity is a CVT.
For relations where the object entity is a CVT, we substitute it with multiple relations using the field types and values of the CVT. Without CVT-based relations, each article has on average 37.67 relations.

\paragraph{WikiText}
The WikiText-derived dataset is constructed using the methods described in Section~\ref{sec:data}. The methods can potentially match overlapping spans, which cannot be handled by NKLM.

Thus, we prune the set of matching spans for each article so that no two spans overlap. Pruning is done by iterating over the spans in a predefined order, and greedily selecting spans that do not overlap with previously selected spans. The spans are ordered by the following criteria:
\begin{itemize}[topsep=4pt,leftmargin=*,itemsep=0pt,parsep=0pt]
  \item In descending order of span length. \textit{(Prefer longer spans)}
  \item In ascending order of span starting index. \textit{(Prefer spans appearing earlier)}
  \item Order spans that match entity canonical forms (the first surface form in list) in front. \textit{(Prefer spans matching canonical forms)}
  \item Ties are broken by relation type ID and index of matched surface form.
\end{itemize}

While NKLM supports partial and arbitrary-order entity matches by specifying copying positions,%
\footnote{For example, the entity ``\textit{Barack Hussein Obama}'' can match the text ``\texttt{Obama Barack}'' with copying positions 2 and 0.}
we do not perform this kind of matching as it greatly increases the complexity of the matching algorithm, and could produce more false positives. We sequentially assign copying positions within matched spans as in WikiFacts.

\section{Comparison of Models using UPP}
\label{app:upp}
We show the main results evaluated according to UPP~\cite{ueberla1994analysing} in Table~\ref{tab:pplupp}.
This adjusted perplexity measure penalizes unknown word probabilities by a constant value of $1 / |\mathcal{V}_{out}|$, where $\mathcal{V}_{out}$ is the set of OOV words in a corpus.

\end{document}